\documentclass[final]{cvpr}

\usepackage{times}
\usepackage{epsfig}
\usepackage{graphicx}
\usepackage{amsmath}
\usepackage{amssymb}
\usepackage{bm} 
\usepackage{booktabs}

\usepackage[pagebackref=true,breaklinks=true,colorlinks,bookmarks=false]{hyperref}

\def\cvprPaperID{****} %
\def\confYear{CVPR 2021}
\begin{document}

\title{Solution for Large-scale Long-tailed Recognition with Noisy Labels}

\author{Yuqiao Xian \\
	Sun Yat-sen University\\ {\tt\small xianyq3@mail2.sysu.edu.cn} \and Jia-Xin Zhuang \\
	Sun Yat-sen University \\ {\tt\small lincolnz9511@gmail.com} \and Fufu Yu \\ Tencent Youtu Lab \\ {\tt\small  fufuyu@tencent.com}%
}

\maketitle

\begin{abstract}
   This is a technical report for CVPR 2021 AliProducts Challenge \footnote{\url{https://tianchi.aliyun.com/competition/entrance/531884/introduction}}. AliProducts Challenge is a competition proposed for studying the large-scale and fine-grained commodity image recognition problem encountered by world-leading e-commerce companies.
   The large-scale product recognition simultaneously meets the challenge of noisy annotations, imbalanced (long-tailed) data distribution and fine-grained classification.
   In our solution, we adopt state-of-the-art model architectures of both CNNs and Transformer, including ResNeSt, EfficientNetV2, and DeiT. We found that iterative data cleaning, classifier weight normalization, high-resolution finetuning, and test time augmentation are key components to improve the performance of training with the noisy and imbalanced dataset. Finally, we obtain 6.4365\% mean class error rate in the leaderboard with our ensemble model. \footnote{This work is finished when Yuqiao Xian and Jia-Xin Zhuang were interns at Tencent Youtu Lab.}
\end{abstract}

\section{Introduction}

Recently, deep learning has shown its impressive performance in computer vision with large well-annotated image datasets such as ImageNet~\cite{ILSVRC15}, MS-COCO~\cite{lin2014microsoft}. However, clean and well-annotated image datasets are hard to obtain and, thus not really in a real-world scenario most time. Datasets like AliProducts~\cite{le2020eCCV}, a noisy and fine-grained product dataset containing 2.5 million web images from the 50,030 fine-grained semantic classes in Stock Keeping Unit (SKU) level, is more challenging and closer to the real-world applications. 

There are several challenges for deep models to achieve high performance in product recognition tasks like AliProducts. Firstly, AliProducts present a long-tail data distribution, whose portions of the distribution haves many occurrences far from the head and central part of the distribution. Secondly, data in AliProducts is very imbalanced and the least number of the category is one of 16 millionths of the largest category. Without any specific strategies for data imbalance, the model would degrade its performance dramatically. Lastly, labels are collected from the internet and are very noisy and data a cleaning algorithm has to be proposed to deal with it~\cite{le2020eCCV}.

To alleviate problems brought by the noisy labels and long-tail distribution in AliProducts, we adopt several strategies to re-balance samples from each class, and then select clean labels in an automatic manner for training models, and train the model with a progressive training scheme. Finally we ensemble CNNs and transformer models to obtain higher accuracy.

\section{Our Solution}

\subsection{Model Architectures}

In our solution, we ensemble three different network architectures with ImageNet pretrained weights, including ResNeSt-101, DeiT-small and EfficientNevV2-m.

\vspace{0.1cm}
\noindent\textbf{ResNeSt} \cite{zhang2020resnest}. ResNeSt is a strong CNN backbone and was adopted by many solutions in last year's competition. We find that ResNeSt-101 can obtain the best performance.

\vspace{0.1cm}
\noindent\textbf{DeiT} \cite{touvron2020deit}. Deit~\cite{touvron2020deit} is selected as transformer branch, which is a kind of vision transformer. We modified DeiT based on  official implementation~\footnote{\url{https://github.com/facebookresearch/deit}}. Consider trade-off between accuracy and efficiency, we use deit-S without distillation from ~\cite{touvron2020deit} in our implementation. We train DeiT with cosine learning for 300 epochs (about 9 days with 8 $\times$ Tesla V100) to get better performance. We find that training DeiT with fewer epochs will result in inferior performance.

\vspace{0.1cm}
\noindent\textbf{EfficientNetV2} \cite{efficientv2}. We find that the EfficientNetV2-m backbone can achieve better validation performance with fewer training epochs. Results of 4 EfficientNetV2-m models with different training settings are included in the final submission.

\vspace{0.1cm}
\noindent\textbf{BNNeck} \cite{Luo_2019_Strong_TMM}. We add a BN layer before the linear classifier to normalize the global features, which slightly improve the performance.

\vspace{0.1cm}
\noindent\textbf{Embedding layer}. Since the number of classes in AliProduct is massvie (50K+), we insert a fully connected layer with 512 output dimensions before the linear classifier to reduce the ites input feature dimensions. It will reduce the parameters of classifier layer $\&$ training time, and will not result in noticable performance drop.

\subsection{Iterative Data Cleaning}

To tackle the severe \textit{label noise} problem in the non-manually labeled training dataset, we iteratively cleaning the training based on model predictions and re-train the model in a self-learning manner. We firstly train the model with the whole training set and use the model to predict the label of training samples. Then, we rank the samples according to the top-1 prediction probability. As shown in Fig. 1, we observe that many samples' predictions are not equal to their labels and also have low prediction probability (confidence). We consider these samples are with noisy labels and abandon them in the next iteration of model training. After 3 - 4 iterations of data cleaning and model re-training, the training dataset will be much cleaner and the model performance can improve about 3\% - 5\% in top-1 accuracy.

\begin{figure}
	\centering
	\vspace{-0.6cm}
	\includegraphics[width=\linewidth]{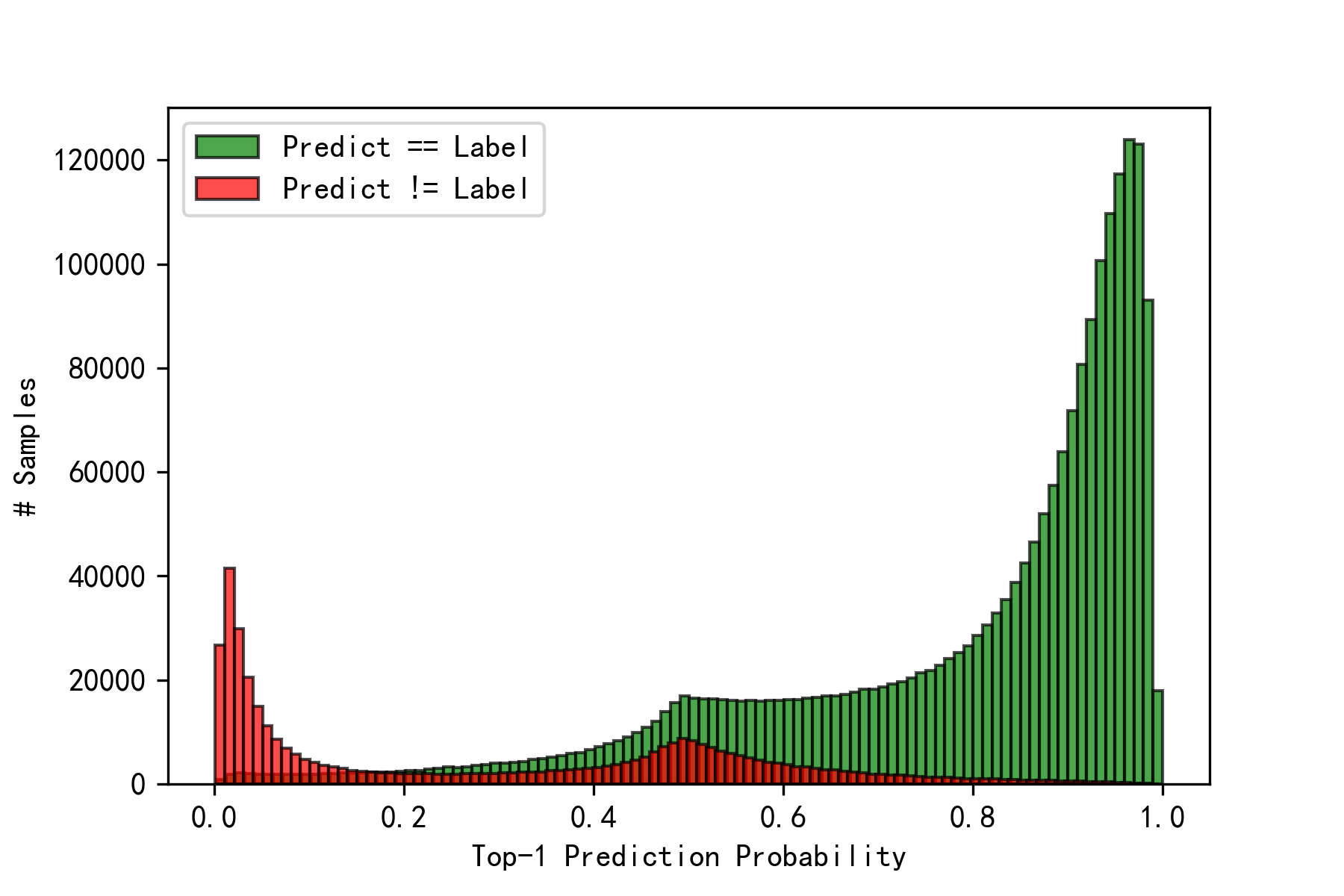}
	\caption{Frequency distribution histogram of model predictions on training samples. Training samples with inconsistent predictions (with their labels) and low prediction probabilities are most likely to be noisy samples.}
	\vspace{-0.2cm}
\end{figure}

\subsection{Training Strategy}

We train all the models using standard cross entropy loss for the classification task, with some strategies to alleviate the negative effect of data imbalance.

\vspace{0.1cm} 
\noindent\textbf{Retraining or Finetuning the Classifier \cite{kang2019decoupling}}. We found that retraining or finetuning the classifier with class-balanced sampling (CBS) is more effective when the training data is noisy, which improve about 3\% - 5\% of the average classification accuracy. After several iterations of data cleaning, the model performance will be higher with instance-balanced sampling (IBS), and finetuning the classifier with CBS improves about 1\% ACC on validation set. Fine-tuning with a balanced training subset is an alternative approach to acquire balanced classifier. We use a trained model to select top 30 images of each class with highest classification probabilities to construct a clean and balance subset and use it to finetune the classifier (or the whole network). We also found these way can improe about 1\% average classification accuray. 

\vspace{0.1cm} \noindent \textbf{High Resolution Finetuning}. Following the idea of progressive training \cite{efficientv2}, we train the ResNeSt-101 and EfficientV2 backbone with $256\times 256$ input size in the first 40 epochs and finetune them with $384 \times 384$ input size for about 5-10 epochs, which can improve about 1\% top-1 accuracy.

\subsection{Test Time Augmentation}

\noindent\textbf{$\bm{\tau}$-Normalization}. \cite{kang2019decoupling} We use $\tau$-normalization to re-balance the decision boundaries of by normalizing the clasifier weights. We found that this method significantly improve the accuracy of tail classes. We found that the optimal value of $\tau$ is different with different model architecture. We choose the value of $\tau$ as 0.5-0.7 based on the validation performance.
\vspace{0.1cm}

\noindent\textbf{Fixing Test Resolution}. Following \cite{touvron2019FixRes}, we scale the resolution of test image by 1.5x of training image. Then, we use ten crop augmentation to crop the enlarged test image to the size of training input. (\textit{e.g.}, $input \rightarrow resized-(256,256) \rightarrow  enlarged-(320, 320) \rightarrow ten cropped-(256,256)$.) This augmentation is quite effective and can improve 1\% - 2\% performance in the most cases.

\vspace{0.1cm}
\noindent\textbf{Model Prediction Ensemble}. To acquire a more robust classification performance, we adaptively ensemble multiple CNN (ResNeSt) and Transformer (DeiT) models with different training settings, such as different dataset versions, different sampling strategies, and different regularizations. We preserve the top-10 predictions of each model for each sample and average the predicting probabilities.

\begin{table}[t]
	\centering 
	\caption{Component analysis on the validation set.}
	\begin{tabular}{l|l}
		\toprule \hline
		Method                       & Top-1 ACC  \\ \hline
		Baseline (EfficientNetV2-m)                   & $\sim0.84$ \\
		+ Data Cleaning              & + $\sim0.02\uparrow\uparrow$ \\
		+ Iterative Cleaning         & +  $\sim0.04\uparrow\uparrow\uparrow$\\
		+ Retraining Classifier      & +  $\sim0.01\uparrow$ \\
		+ $\tau$-Normalization       & +  $\sim0.02\uparrow\uparrow$ \\
		+ TTA (Enlarge + Ten Crop)   & +  $\sim0.02\uparrow\uparrow$ \\
		+ High Resolution Finetuning & +  $\sim0.01\uparrow$ \\
		+ 3 Model Ensemble           & +  $\sim0.01\uparrow$ \\
		+ 8 Model Ensemble           & +  $\sim0.02\uparrow\uparrow$ \\ \hline
	\end{tabular}
\vspace{-0.2cm}
\end{table}

\subsection{Other Techniques}

We also use the following techniques to train some of our models, which may slightly improve the performance on the validation set. These models are included in the final ensemble model. However, due to the limited chances of submission, we are not able to provide a comprehensive ablation study and evaluate each component on test set:
1. HAR \cite{cao2020heteroskedastic};
2. Large weight decay rate (0.0005);
3. Mixup;
4. Relabeling the filtered training Wimages with model prediction.

We also implement the following techniques, but can not find a way the improve the validation performance with them: 
1. ArcFace \cite{deng2018arcface}. 2. Learnable weight scaling \cite{kang2019decoupling}. 3. CutMix \cite{yun2019cutmix}. 4. Nearest Class Mean classifier \cite{kang2019decoupling}.

\section{Implementation Settings}
Most of our models are trained on a node with 8 NVIDIA Tesla V100 GPUs on Tencent Jizhi Computation Platform. A typical EfficientV2-m model is trained with 30 epochs with IBS and other 10 epochs with CBS using standard SGD optimizer with momentum, which needs about 30 hours. The settings are as following:

\begin{itemize}
	\item Data Augementation: Resize to $256\times 256$ with random crop, random hue saturation shift, random brightness contrast shift, random gaussian blur, random horizontal flip, random vertical flip and coarse dropout.
	\item Step learning rate decay at 14 and 20 epoch (ratio 0.1).
	\item Warmup training at the first 2 epochs.
	\item Initial learning rate of $0.01 \times \frac{batch\_size}{256}$
	\item Label smoothing: 0, 0.1 and 0.2.
	\item Batch size: 1024.
	
	\item Dimensions of embedding layer: 512.
	\item Weight decay: $0.0002 - 0.0005$.
\end{itemize}

For transformer, a typical training of 300 epochs takes 9 days with 8 V100 for the deit-S. We follow the default settings~\cite{touvron2020deit} in most time. Learning rate is 5e-4 with cosine scheduler with AdamW. A batch contains 2048 images. Images are resized to $256\times 256$ and random cropped by $224\times 224$. Many data augmentations methods such as random augment, repeated augmentation is used to train our model. Note that, we follow the default strategies to train deit~\cite{touvron2020deit} and use repeated augmentations with 3 repetitions. %

\section{Conclusion}

Since AliProducts presents noisy labels and long-tail distribution, we select clean images based on confidence score from the 	model prediction and then adopt several effective strategies to re-balance samples from each class and train the model with strong regularization terms in a progressive way. Ensemble of strong yet diverse backbones like CNNs and Transformer would further improve generalization and boost the performance. Finally, we achieve 0.064365 mean class error rate in the leaderboard with our solution.

\section*{Acknowledgement}
Gratitude to CVPR 2021 organizing Committee and Alibaba TianChi platform for organizing this competition. Tencent Youtu Lab and Tencent Jizhi Computation Platform provide the technical and computation support for our solution. Thanks to Enwei Zhang and Xing Sun from Tencent Youtu Lab to provide precious opinions to our solution. 

{\small
\bibliographystyle{ieee_fullname}
\bibliography{egbib}

\begin{thebibliography}{10}\itemsep=-1pt

\bibitem{cao2020heteroskedastic}
Kaidi Cao, Yining Chen, Junwei Lu, Nikos Arechiga, Adrien Gaidon, and Tengyu
  Ma.
\newblock Heteroskedastic and imbalanced deep learning with adaptive
  regularization.
\newblock In {\em Int. Conf. Learn. Represent.}, 2021.

\bibitem{le2020eCCV}
Lele Cheng, Xiangzeng Zhou, Liming Zhao, Dangwei Li, Hong Shang, Yun Zheng, Pan
  Pan, and Yinghui Xu.
\newblock Weakly supervised learning with side information for noisy labeled
  images.
\newblock In {\em Eur. Conf. Comput. Vis.}, 2020.

\bibitem{deng2018arcface}
Jiankang Deng, Jia Guo, Xue Niannan, and Stefanos Zafeiriou.
\newblock Arcface: Additive angular margin loss for deep face recognition.
\newblock In {\em IEEE Conf. Comput. Vis. Pattern Recog.}, 2019.

\bibitem{kang2019decoupling}
Bingyi Kang, Saining Xie, Marcus Rohrbach, Zhicheng Yan, Albert Gordo, Jiashi
  Feng, and Yannis Kalantidis.
\newblock Decoupling representation and classifier for long-tailed recognition.
\newblock In {\em Int. Conf. Learn. Represent.}, 2020.

\bibitem{lin2014microsoft}
Tsung-Yi Lin, Michael Maire, Serge Belongie, James Hays, Pietro Perona, Deva
  Ramanan, Piotr Doll{\'a}r, and C~Lawrence Zitnick.
\newblock Microsoft coco: Common objects in context.
\newblock In {\em Eur. Conf. Comput. Vis.}, pages 740--755.

\bibitem{Luo_2019_Strong_TMM}
H. {Luo}, W. {Jiang}, Y. {Gu}, F. {Liu}, X. {Liao}, S. {Lai}, and J. {Gu}.
\newblock A strong baseline and batch normalization neck for deep person
  re-identification.
\newblock {\em IEEE Trans. Multimedia}, 2019.

\bibitem{ILSVRC15}
Olga Russakovsky, Jia Deng, Hao Su, Jonathan Krause, Sanjeev Satheesh, Sean Ma,
  Zhiheng Huang, Andrej Karpathy, Aditya Khosla, Michael Bernstein,
  Alexander~C. Berg, and Li Fei-Fei.
\newblock {ImageNet Large Scale Visual Recognition Challenge}.
\newblock In {\em Int. J. Comput. Vis.}, 2015.

\bibitem{efficientv2}
Mingxing Tan and Quoc~V. Le.
\newblock Efficientnetv2: Smaller models and faster training.
\newblock 2021.

\bibitem{touvron2020deit}
Hugo Touvron, Matthieu Cord, Matthijs Douze, Francisco Massa, Alexandre
  Sablayrolles, and Herv\'e J\'egou.
\newblock Training data-efficient image transformers \& distillation through
  attention.
\newblock {\em arXiv preprint arXiv:2012.12877}, 2020.

\bibitem{touvron2019FixRes}
Hugo Touvron, Andrea Vedaldi, Matthijs Douze, and Herv{\'e} J{\'e}gou.
\newblock Fixing the train-test resolution discrepancy.
\newblock In {\em Adv. Neural Inform. Process. Syst.}, 2019.

\bibitem{yun2019cutmix}
Sangdoo Yun, Dongyoon Han, Seong~Joon Oh, Sanghyuk Chun, Junsuk Choe, and
  Youngjoon Yoo.
\newblock Cutmix: Regularization strategy to train strong classifiers with
  localizable features.
\newblock In {\em Int. Conf. Comput. Vis.}, 2019.

\bibitem{zhang2020resnest}
Hang Zhang, Chongruo Wu, Zhongyue Zhang, Yi Zhu, Zhi Zhang, Haibin Lin, Yue
  Sun, Tong He, Jonas Muller, R. Manmatha, Mu Li, and Alexander Smola.
\newblock Resnest: Split-attention networks.
\newblock {\em arXiv preprint arXiv:2004.08955}, 2020.

\end{thebibliography}
}

\end{document}